\pdfoutput=1
\documentclass[sigconf,natbib=true,anonymous=false]{acmart}

\AtBeginDocument{%
  }

\copyrightyear{2023} 
\acmYear{2023} 
\setcopyright{acmlicensed}\acmConference[SIGIR '23]{Proceedings of the 46th International ACM SIGIR Conference on Research and Development in Information Retrieval}{July 23--27, 2023}{Taipei, Taiwan}
\acmBooktitle{Proceedings of the 46th International ACM SIGIR Conference on Research and Development in Information Retrieval (SIGIR '23), July 23--27, 2023, Taipei, Taiwan}
\acmPrice{15.00}
\acmDOI{10.1145/3539618.3591911}
\acmISBN{978-1-4503-9408-6/23/07}

\usepackage{microtype}
\usepackage{float}
\usepackage{multirow}
\usepackage{balance}

\begin{document}

\title{REFinD: Relation Extraction Financial Dataset}

\author{Simerjot Kaur$^*$}
\email{simerjot.kaur@jpmchase.com}
\affiliation{
\institution{JPMorgan Chase and Co}
\city{Palo Alto}
\state{CA}
\country{USA}}

\author{Charese Smiley$^*$}
\email{charese.h.smiley@jpmchase.com}
\affiliation{
\institution{JPMorgan Chase and Co}
\city{Chicago}
\state{IL}
\country{USA}}

\author{Akshat Gupta}
\email{akshat.x.gupta@jpmchase.com}
\affiliation{
\institution{JPMorgan Chase and Co}
\city{New York}
\state{NY}
\country{USA}}

\author{Joy Sain}
\email{sain.9@wright.edu}
\affiliation{
\institution{Wright State University}
\city{Dayton}
\state{OH}
\country{USA}}

\author{Dongsheng Wang}
\email{dongsheng.wang@jpmchase.com}
\affiliation{
\institution{JPMorgan Chase and Co}
\city{London}
\country{UK}}

\author{Suchetha Siddagangappa}
\email{suchetha.siddagangappa@jpmchase.com}
\affiliation{
\institution{JPMorgan Chase and Co}
\city{New York}
\state{NY}
\country{USA}}

\author{Toyin Aguda}
\email{toyin.d.aguda@jpmchase.com}
\affiliation{
\institution{JPMorgan Chase and Co}
\city{Chicago}
\state{IL}
\country{USA}}

\author{Sameena Shah}
\email{sameena.shah@jpmchase.com}
\affiliation{
\institution{JPMorgan Chase and Co}
\city{New York}
\state{NY}
\country{USA}}

\renewcommand{\shortauthors}{Simerjot Kaur et al.}


\begin{abstract}
  A number of datasets for Relation Extraction (RE) have been created to aide downstream tasks such as information retrieval, semantic search, question answering and textual entailment. However, these datasets fail to capture financial-domain specific challenges since most of these datasets are compiled using general knowledge sources, hindering real-life progress and adoption within the financial world. To address this limitation, we propose REFinD, the first large-scale annotated dataset of relations, with $\sim$29K instances and 22 relations amongst 8 types of entity pairs, generated entirely over financial documents. We also provide an empirical evaluation with various state-of-the-art models as benchmarks for the RE task and highlight the challenges posed by our dataset. We observed that various state-of-the-art deep learning models struggle with numeric inference, relational and directional ambiguity. 
\end{abstract}





\maketitle

\section{Introduction} \label{intro}

\begin{figure*}[tp]
\centering
\includegraphics[width=15.5cm,height=3cm]{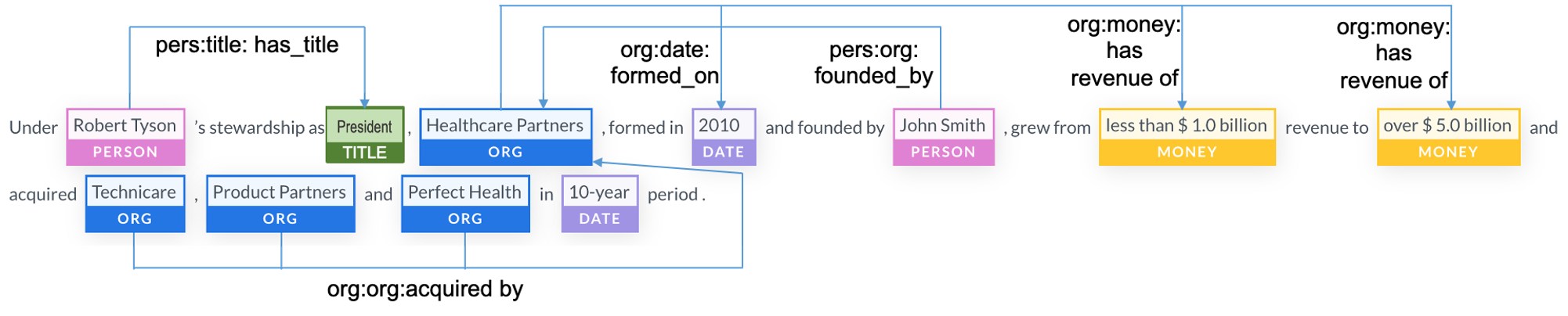}
\caption{Example sentence from a financial report with highlighted entities and relations. In a single sentence, there are 8 relations present: \textit{has title}, \textit{formed in}, \textit{founded by}, 2 instances of \textit{has revenue of}, and 3 instances of \textit{acquired by}.} 
\label{fig:figeg}
\end{figure*}

The exponential progress of AI across multiple domains can largely be attributed to the availability of large datasets coupled with an increase in available compute power.  Relation extraction (RE) from text is a fundamental problem in NLP and information retrieval, which facilitates various tasks like knowledge graph construction, question answering and semantic search. It has seen significant progress in recent years, thanks to advanced machine learning techniques and the availability of large-scale relation extraction datasets. However, most existing large-scale RE datasets are derived from general knowledge sources such as Wikipedia, web texts and news articles \cite{hendrickx-etal-2010-semeval, zhang2015relation, zhang2017position, sharma2022finred, jabbari2020french}. These datasets often fall short in addressing domain-specific challenges. Hence, various state-of-the-art models that perform competitively on such datasets fail to perform well in the financial domain (shown in Section \ref{sec:benchmark_exp}). In particular, financial text documents, such as financial reports and various Securities and Exchange Commission (SEC) filings, differ significantly from standard English language documents. They necessitate the extraction of entities and relations that involve numbers, currencies, dates, legal facts, and claims, often embedded in longer and more complex sentences with substantial distances between entities. Figure 1 illustrates a prototypical sentence from a financial report, emphasizing unique relations like \textit{acquired by}, \textit{revenue of}. 

Moreover, financial documents often necessitate more advanced numerical inference to identify relationships amongst entities. For instance, a company is deemed to have been acquired by another entity if the latter owns more than 50\% of its shares. Additionally, there can be ambiguity amongst relations in financial text, such as when a person serves only on a company's board is just a member of company and not considered as an employee. These financial domain-specific challenges make relation extraction from such documents more difficult. However, the current absence of large-scale finance-specific relation extraction dataset impedes progress, benchmarking, and real-life adoption of various relation extraction algorithms within the financial industry. 

\def\thefootnote{*}\footnotetext{These authors contributed equally to this work}\def\thefootnote{\arabic{footnote}}


To address this limitation, we have developed the largest relation extraction dataset for financial documents to date, REFinD \footnote{\url{https://www.jpmorgan.com/technology/artificial-intelligence/initiatives/refind-dataset/problem-motivation-outcome}}, which contains $\sim$29K instances and 22 relations among 8 types of entity pairs. REFinD is a domain-specific financial relation extraction dataset created using raw text from various 10-X reports (including 10-K, 10-Q, etc. broadly known as 10-X) of publicly traded companies obtained from US Securities and Exchange Commission (SEC)\footnote{https://www.sec.gov/edgar.shtml} 
website (detailed in Section \ref{sec:data_construction}). Although primarily built on financial reports and focused on financial domain-specific challenges, this dataset can also be leveraged by other domains such as legal, risk modeling, and econometrics. 

In this work, we also highlight various challenges associated with creating a large-scale relation extraction dataset specifically over financial-domain. Since financial documents contain much longer and complex sentences and inferring relations from them involves a tremendous amount of financial-domain expertise, hence collecting seed labels, removing noisy text, and finally annotating such dataset becomes an extremely challenging task. Finally, we also provide benchmarks on REFinD dataset to identify and highlight challenges it poses, as well as to spur further research and improvements in the field of financial relation extraction. We observed that despite fine-tuning various state-of-the-art deep learning models on the REFinD dataset, their performance remains sub-optimal on finance-specific relations. Even specialized models like FinBERT and FLANG, which have been further trained on financial news articles and incorporate masking for financial-terms, do not demonstrate significant improvement. This is likely due to their lack of exposure to semantically complex financial documents, which hinders their ability to effectively handle intricate finance-specific scenarios.

This resource paper presents the following key contributions:
\begin{itemize}
    \item Introducing REFinD, the first large-scale Relation Extraction Dataset over financial documents.
    \item Establishing benchmarks for various state-of-the-art models using the REFinD dataset.
    \item Identifying and highlighting the unique financial-domain specific challenges posed by the REFinD dataset.
\end{itemize}


\section{Related Work} \label{sec:related_work}


Several datasets have been developed for RE using general knowledge sources such as Wikipedia\footnote{https://www.wikipedia.org}  and web articles including ACE 2003-2004 \cite{strassel2008linguistic}, SemEval\-2010 Task 8 \cite{hendrickx-etal-2010-semeval}, KBP37 \cite{zhang2015relation}, TACRED \cite{zhang2017position}, FewRel \cite{han2018fewrel}, CrossRE \cite{bassignana2022crossre}, and MAVEN-ERE \cite{wang2022maven}. However, financial texts pose their own unique set of challenges and there has been limited attention paid towards creating RE datasets within the financial domain. Recently, a few datasets have been developed using financial news and earnings calls, including FinRED \cite{sharma2022finred}, CorpusFR \cite{jabbari2020french} and Financial News Corpus \cite{Wu2020CreatingAL}. Table \ref{tab:dataset_comparison} compares the number of relations and instances for recent general-purpose datasets (top) and financial datasets (bottom).

\begin{table}[h!]
\begin{tabular}{lll}
\hline
Dataset & Rels & Instances \\ \hline
SemEval-2010 Task 8 & 19 & 10,717 \\
TACRED & 42 & 119,474 \\
KBP37 & 37 & 21,046 \\ \hline
FinRED & 29 & 6,767 \\
Financial News Corpus & 6 & 22,812 \\
CorpusFR & 20 & 1,754 \\
REFinD & 22 & 28,676 \\ \hline
\end{tabular}
\caption{Comparison between REFinD and prior RE datasets.  REFinD is a large dataset, second only to TACRED, although it targets fewer relations.}  
\label{tab:dataset_comparison}
\end{table}

\begin{figure}[b!]
    \centering
    \includegraphics[width=\columnwidth]{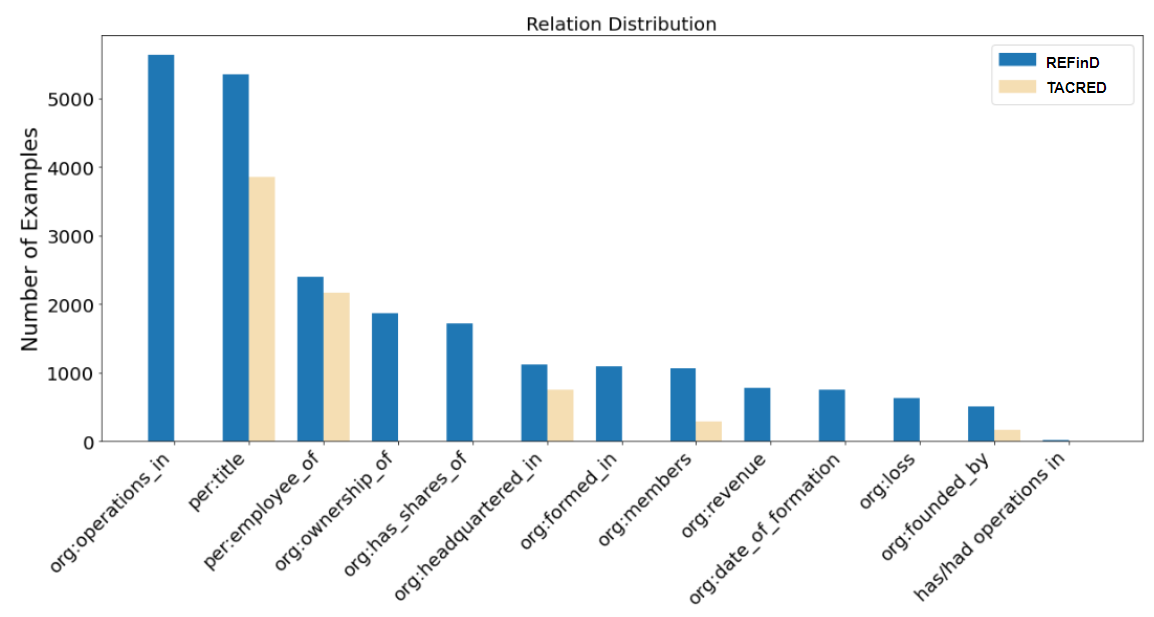}
    \caption{Relationship Distribution in REFinD and TACRED datasets}
    \label{fig:reldist} 
\end{figure}

Table \ref{tab:dataset_comparison} shows that TACRED, which annotates 42 relations, is the largest of these datasets. However, it should be noted that 79.5\% of the TACRED dataset comprises \textsc{no\_relation} instances, whereas only 45.5\% of REFinD is \textsc{no\_relation}. Consequently, while REFinD has a smaller total number of relations than TACRED, it has a higher number of instances for each relation it covers. Moreover, REFinD has greater coverage than TACRED for the relations of interest in finance, as shown in Figure \ref{fig:reldist}. Among financial datasets, both FinRED and CorpusFR are small compared to REFinD, and Financial News Corpus covers fewer relation types. Finally, REFinD is the first RE dataset to utilize SEC filings, which are a rich and complex data source.

\section{Dataset Overview and Construction} \label{sec:data_construction}

\begin{figure}[h!]
\begin{center}
\centerline{\includegraphics[width=0.9\columnwidth]{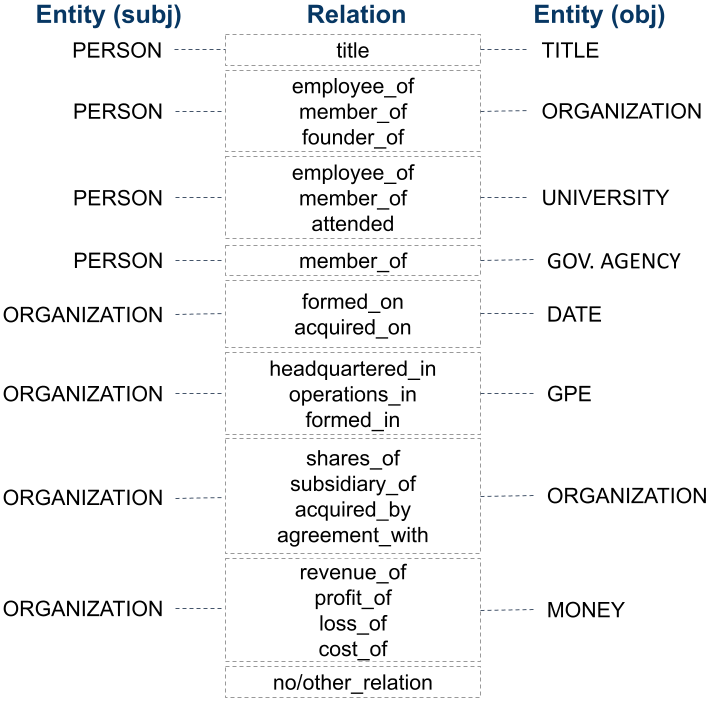}}
\caption{\textbf{REFinD Dataset}. The dataset has 8 Entity pair groups and 22 Relation types.} 
\label{fig:RelationGroup}
\end{center}
\end{figure}

REFinD is specifically designed for use within the financial domain, using 10-X filings obtained from the SEC. The dataset targets 8 finance-specific entity pairs:  \textsc{person-title}, \textsc{person-org}, \textsc{person-univ}, \textsc{person-gov\_age-} \textsc{ncy}, \textsc{org-gpe}, \textsc{org-date}, \textsc{org-org}, and \textsc{org-money}. Each entity pair group includes several possible finance-oriented relation types and Figure \ref{fig:RelationGroup} illustrates these entity pair groups along with the 22 proposed relation types in REFinD dataset. 

Our annotations are at the instance-level, and each instance corresponds to the directed entity pair in a sentence that is annotated with one of the 22 relation labels. The dataset statistics and its distribution are provided in Table \ref{tab:relation_breakdown} (Appendix \ref{refinddatastats}). Additionally, we have included a snippet of the REFinD dataset in Appendix \ref{app:datasnip}. In total, the REFinD dataset contains $\sim$29K instances with an average sentence length of 53 words and average contextual complexity of 11 words between the entity pairs.  

The following sections provide detailed information on the collection process of the REFinD dataset, as well as the preprocessing steps that were taken to obtain high-quality data. Additionally, the annotation process used to label the instances with the 22 proposed relation types is explained in detail.

\subsection{Document Collection}
While most RE datasets have been constructed using sources such as Wikipedia, web articles, and financial news articles \cite{hendrickx-etal-2010-semeval, zhang2015relation, zhang2017position, sharma2022finred, jabbari2020french}, the REFinD dataset is designed specifically for use within the financial domain and has been constructed from 10-X filings downloaded from the SEC website for the years 2016-2017. These publicly available financial regulatory reports are submitted to the SEC at regular intervals (monthly/quarterly/annually) by publicly traded companies and issuers of securities. They often provide detailed information about ownership, executive compensation, corporate structure, shares, trades, and other financial details.

\subsection{Preprocessing}
To prepare the 10-X reports for annotation, extensive data cleaning was required to address noise such as HTML tags and redundant spaces. Additionally, specific financial terms used to refer to legal entities presented a challenge, requiring entity tagging at the sentence-level to build the knowledge base.

\noindent
\textbf{Data Cleaning:} To prepare the corpus for annotation, we first performed extensive data cleaning to remove header and footer information, tables, HTML tags, and redundant spaces. Additionally, financial text has unique characteristics, including the use of specific financial terms to refer to legal entities. To resolve these terms, we replaced pronouns and referring terms such as `we', `our', and `the company' with the corresponding organization name. For example, the sentence such as \textit{\underline{The Company} had net revenues of \$14,720,545}, with organization name as \textit{Technicare, LLC}, is preprocessed as \textit{\underline{Technicare, LLC} had net revenues of \$14,720,545}.

\noindent
\textbf{Named Entity Tagging:} In order to build the knowledge base, we performed sentence tokenization and part-of-speech (POS) tagging using the spaCy library \cite{spacy}. Named entity recognition (NER) was also carried out using spaCy for five entity types: \textsc{per}, \textsc{org}, \textsc{date}, \textsc{gpe}, and \textsc{money}. To capture the extensive use of job titles within financial documents, we employed Stanford CoreNLP \cite{manning2014stanford} to tag entities as \textsc{title}. Additionally, we introduced two new entity types: \textsc{univ} for educational establishments such as schools, colleges, universities, and institutes, and \textsc{gov} for U.S. government agencies, using Gazetteer\footnote{\url{https://census.gov}} lists and regular expressions, respectively. 

\subsection{Dataset Construction}
In order to construct this dataset, $\sim$26K filings per year (2016-2017) were downloaded and after extracting and preprocessing we are left with millions of instances. Manual annotation of each instance is prohibitively costly and due to the sparsity of relations in the dataset (e.g. the location of a company's headquarters may only be mentioned once per document), using a random sample of instances for annotation would lead to an extremely imbalanced dataset.  Moreover, distantly supervised techniques such as \citet{mintz-etal-2009-distant} work well for generating datasets, but rely on the use of an external knowledge base (KB) whereas financial text contains relatively unknown person/legal entities and finance-oriented relations which cannot be tracked by a general purpose KB.  

Hence, to achieve purposeful annotation, we utilized a context-sensitive approach based on the construction of a set of phrases \cite{zuo2017uncovering,agichtein2000snowball}.  Specifically, we focused on pairs of entities $e_1$ and $e_2$ and a set of seed phrases $p$ that were relevant to each relation type. For instance, we used phrases such as `net revenue' for \textsc{org-money} and `served as' for \textsc{per-org}. We retained instances that contained these phrases and collected all patterns $(e_1,e_2,p)$ within the target relation group. We then used the \textsc{Hit} score, as in \citet{zuo2017uncovering}, to rank these patterns and retained the top $k$ patterns. We then calculate the convergence score \textit{Conv} to filter out candidates that co-occur with few entity pairs and retain those with a threshold greater than $\tau$ (see Appendix Eq. (\ref{hit-score})).

\begin{figure}
    \centering
    \includegraphics[width=\columnwidth]{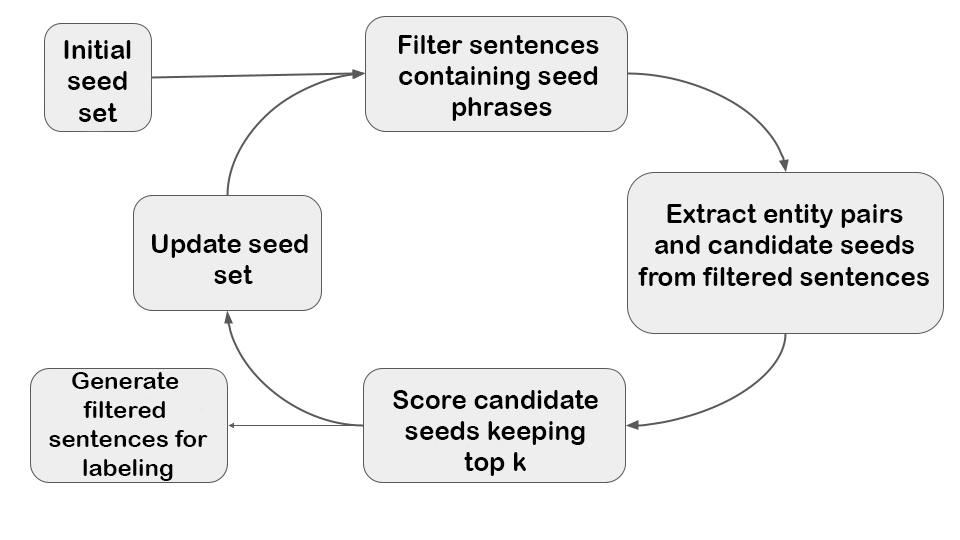}
    \caption{Corpus filtering using initial seed set.}
    \label{fig:sentence_filtering}
\end{figure}


Finally, to introduce diversity in our dataset, we expanded upon the approach used in \cite{zuo2017uncovering} by collecting phrases from the shortest dependency path (SDP) between $e_1$ and $e_2$. This captures longer contexts seen in financial texts and includes all the context surrounding the entities. We then created a new set of candidate seeds and retained the top $k$ based on their \textsc{Hit} and \textit{Conv} scores. The selected seeds were then used in the next round of filtering, as shown in Figure \ref{fig:sentence_filtering}. The process was repeated until no new seeds were added to the list or a maximum number of cycles was reached. The expanded seed list was then used to filter the instances and obtain the final set for annotation. This final set contained a more concentrated number of target relations, introduced more diverse expressions, and included near-misses that should be categorized as \textit{no\_relation}.

\subsection{Dataset Annotation}
To annotate our constructed dataset, we leverage Amazon Mechanical Turk (MTurk)\footnote{\url{https://www.mturk.com}}, a platform that allows us to crowdsource human intelligence tasks. Our dataset annotation task involves presenting a Human Intelligence Task (HIT) to a crowdworker. In each HIT, two entities, $e_1$ and $e_2$, are highlighted, and the worker is asked to select the relation $r$ that best describes the ordered entity pair ($e_1$, $e_2$) from a list of relations presented to them (see Figure \ref{fig:RelationGroup}). The list includes all possible entity pair relations, as well as an option for \textit{no\_relation / different\_relation}. To ensure the quality of the annotations, we provide clear instructions and examples of how to complete the task, an example of how the annotation task is presented to the crowdworker can be found in Appendix \ref{mturkeg}.


\subsection{Annotation Guidelines}

To ensure high-quality annotations, we developed guidelines for the annotation task through an iterative process. This involved two preliminary rounds on a subset of data, followed by two official rounds for the 2016 data, one official round for the 2017 data, and one round for both years, all of which were reviewed by financial experts. During the preliminary rounds, we observed that workers experienced ambiguity when it came to the temporal aspect of relations in some instances. To address this, we provided a list of relations for each instance, along with the present and past tense verbs. For example, to show the relation between a \textsc{person} and an \textsc{org}, we listed the relation as \textit{$e_1$ is/was an employee of $e_2$}, where entity markers $e_1$ and $e_2$ are replaced with the relevant entities. 

\begin{table}[h]
\begin{tabular}{|c|c|c|}
\hline
\textbf{Entity Group} & \multicolumn{1}{l|}{\textbf{Original Relation}}                   & \textbf{Modified Relation}                                                                                         \\ \hline
PER:ORG               & \begin{tabular}[c]{@{}c@{}}is/was an \\ employee of\end{tabular}  & \begin{tabular}[c]{@{}c@{}}is/was an employee of \\ (e.g. CEO, President, \\ Vice President, Manager)\end{tabular} \\
PER:ORG               & \begin{tabular}[c]{@{}c@{}}is/was a \\ member of\end{tabular}     & \begin{tabular}[c]{@{}c@{}}is/was a (board, \\ committee) member of\end{tabular}                          \\
PER:TITLE             & \begin{tabular}[c]{@{}c@{}}has/had \\ job title\end{tabular}      & \begin{tabular}[c]{@{}c@{}}has/had job title \\ (e,g. CEO, manager, \\ director)\end{tabular}                      \\
ORG:GPE               & \begin{tabular}[c]{@{}c@{}}has/had \\ operations in\end{tabular}  & \begin{tabular}[c]{@{}c@{}}has/had operations in \\ (headquartered \\ elsewhere)\end{tabular}                  \\
ORG:ORG               & \begin{tabular}[c]{@{}c@{}}has/had \\ shares of\end{tabular}      & \begin{tabular}[c]{@{}c@{}}has/had \% or \\ number of shares of\end{tabular}                                       \\ \hline
\end{tabular}
\caption{Relation labels modified during annotation rounds to enhance accuracy of annotations.}
\label{tab:guidelines} 
\end{table}

We also found that some workers would select the \textit{founder\_of} relation for any high-ranking position, such as the CEO or President, in cases where the roles of \textit{founder\_of} and \textit{employee\_of} may not necessarily overlap. To address this, we specified in our guidelines that workers should choose the relation that is most clearly stated in the displayed instance. Additionally, to increase clarity, we included examples in the relation text shown to workers. We modified the list of relations, and a complete list is provided in Table \ref{tab:guidelines}. Overall, our guidelines aim to reduce ambiguity and ensure consistent and accurate annotations.

For the 2016 data, we conducted two rounds of annotations during the official rounds. In the first round, we collected two judgements for all instances. For the second round, we collected one more judgement, but only for the instances where there was no consensus in the first round. However, we realized that this approach was time-consuming and required multiple rounds of labeling. To improve the efficiency of the annotated data collection process, for the 2017 data, we increased the number of judgements collected in the first round. We collected the number of judgements equal to the number of options within an entity pair group. This approach was more expensive, but it helped us save time by reducing the number of rounds needed for labeling. After collecting all the annotations from MTurk, we conducted one final official round. This round involved adjudication by experts in the finance domain to ensure that all the judgements were accurate and reliable.

\section{Dataset Analysis and Quality}
\label{sec:data_analysis}

\subsection{Contextual Complexity}

\begin{figure}[h!]
    \centering
    \includegraphics[width=\columnwidth]{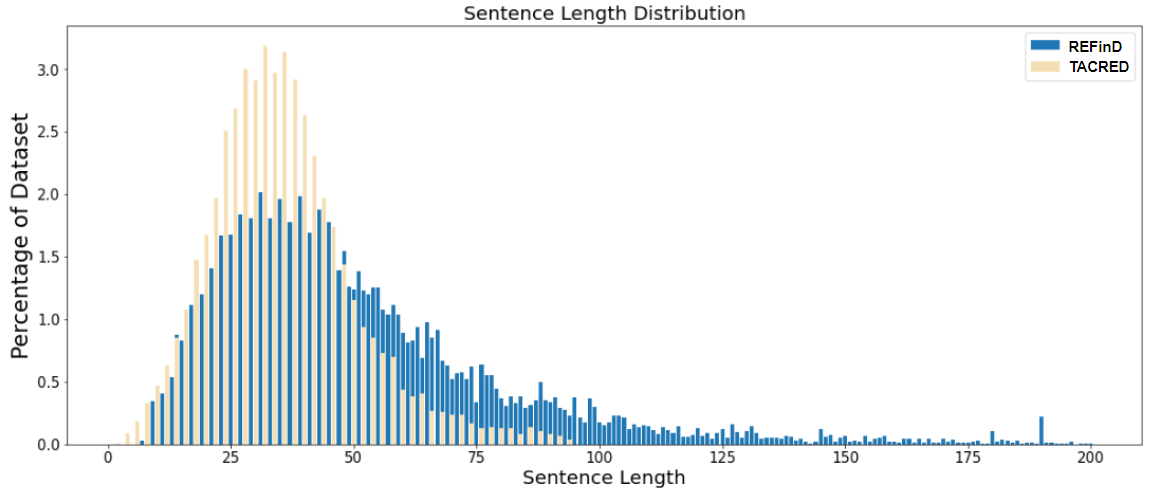}
    \caption{A detailed comparison of the sentence length distributions between REFinD and TACRED datasets.}
    \label{fig:sentlendist}
\end{figure}

To evaluate the contextual complexity of financial-domain instances, we compared the sentence length distributions for REFinD and the largest known Wikipedia relations dataset, TACRED. As illustrated in Figure \ref{fig:sentlendist}, our analysis reveals that REFinD contains much longer sentences than TACRED wherein the average sentence length in TACRED dataset is 36.2 while REFinD is 53.7. 

\begin{figure}[h!]
    \centering
    \includegraphics[width=1.0\columnwidth]{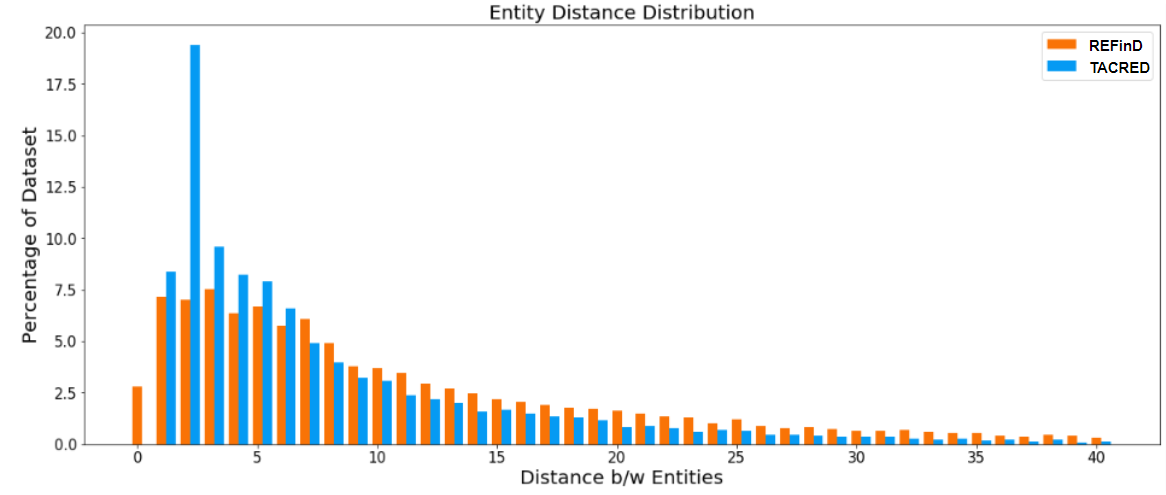}
    \caption{A detailed comparison of the contextual complexity between REFinD and TACRED datasets.}
    \label{fig:entitylendist}
\end{figure}

We also compared the distance between entity pairs in REFinD and TACRED. As illustrated in Figure \ref{fig:entitylendist}, our analysis indicates that REFinD includes more complex sentences than TACRED, with an average entity-pair distance of 11, compared to 8 in TACRED. This finding suggests that REFinD presents a greater level of difficulty in terms of identifying and linking entities within sentences, further emphasizing the dataset's contextual complexity in the financial domain.

\subsection{Label Aggregation} \label{labagg}

Our data collection efforts resulted in a total of 28,676 assessments from 1,209 assessors. To ensure the accuracy and reliability of the annotations, we aggregated the assessments using an internally calculated trust score that takes into account each assessor's performance. 

To calculate the trust score score for each assessor as follows: for each assessor $w$, we build a $|L| \times |L|$ confusion matrix $F$ such that the $(i,j)$th entry is the count of times the assessor assigned label $i$ to a label $j$. Thus, the accuracy is simply the sum of the diagonal entries divided by the total number of assessments made by the assessor. However, this approach leads to a potential bias against the assessor, because an assessor could have used a label close to, but not identical to the ground truth label, e.g., \textit{founder\_of} versus \textit{employee\_of}. Therefore, to address the issue, we multiply the performance confusion matrix by a label similarity matrix $Sim$ of size $|L| \times |L|$. This yields an adjusted confusion matrix $F^* = F \times Sim$, which takes into account the similarity between labels. While the label similarity matrix can theoretically take any form, we typically collaborate with domain experts who have reviewed the instances in the dataset to generate it. Then, we define the reliability of assessor $trust_w$ as the sum of all values of the diagonal entries of $F^*$. 

\begin{figure}[h!]
    \includegraphics[width=0.7\columnwidth]{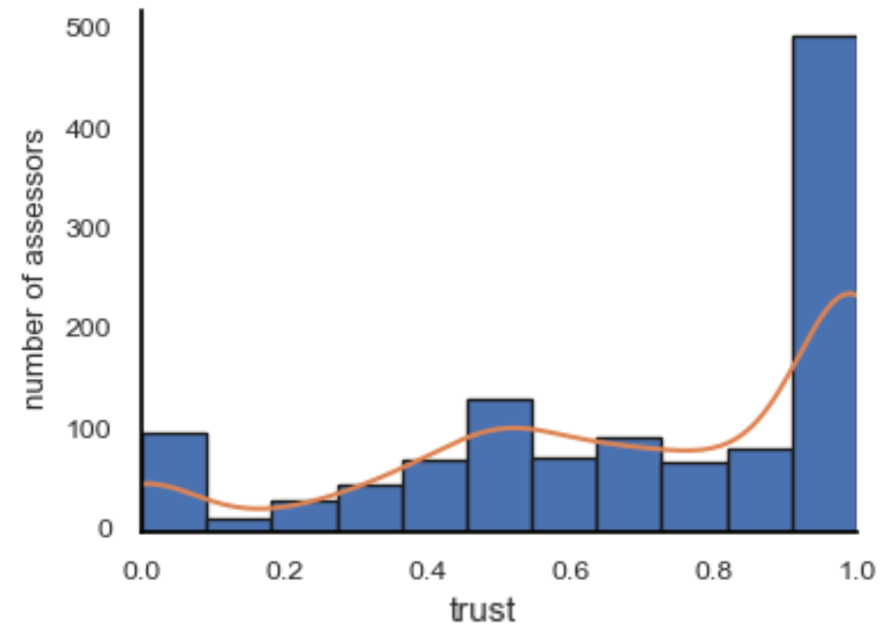}
    \caption{The trust distribution of the assessors, x-axis is the trust score $trust_w$ between 0 and 1.}
    \label{fig:workertrustdist}
\end{figure}

Figure \ref{fig:workertrustdist} shows the distribution of trust levels among assessors who annotated more than three instances. Out of a total of 1,209 assessors, 1,116 met this criterion. The range of trust levels is from 0 to 1, indicating variability in assessor performance. Some assessors may be spammers or lack sufficient domain knowledge to contribute effectively to the task. It's also worth noting that assessors with lower trust levels may have assessed fewer instances.

\begin{figure*}[btp]
    \centering
\includegraphics[width=1.6\columnwidth]{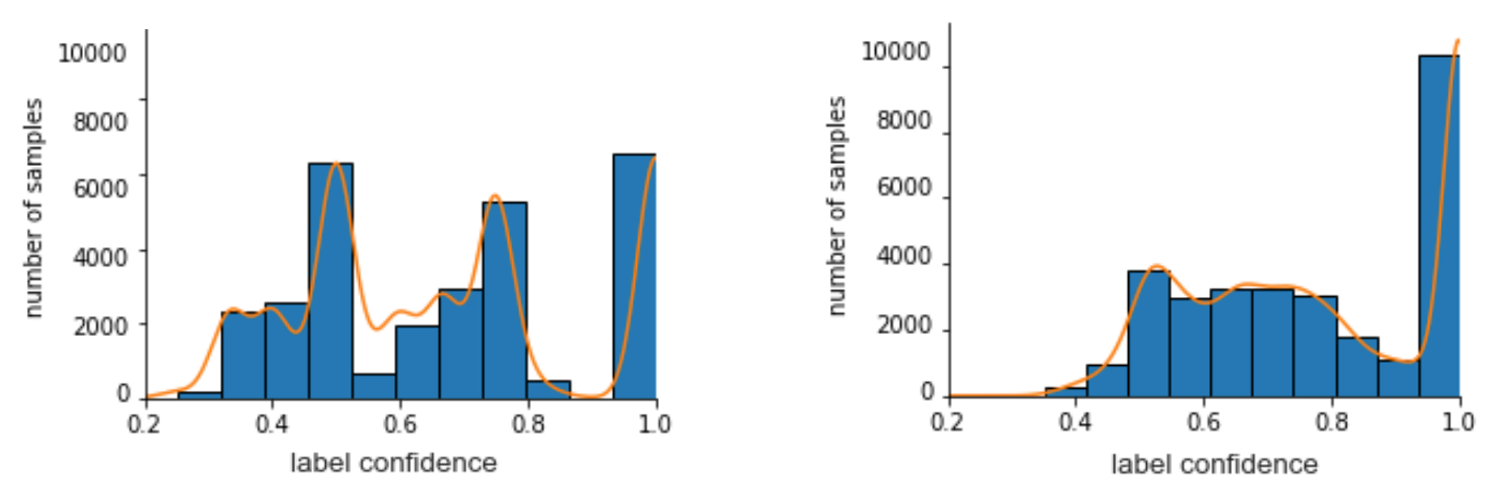}
    \caption{The normalized aggregated confidence for all samples (i) before expert checking and (ii) after expert checking, where x-axis is the confidence score between 0 and 1.}
    \label{fig:ass}
\end{figure*}

Thus, the vote for an instance with label $l$ is formulated with the consideration of reliability as,
\begin{equation} \label{voteeq}
    \text{vote}(l) = \sum_{w \in W} trust_w(l)
\end{equation}

Finally, for a given instance and pair of two entities, we assign the label with the maximum vote score,
\begin{equation}
    (s,e_1,e_2) \mapsto \bar{l} \text{ where} \quad \bar{l} = \text{argmax}_{l \in L} \text{vote(l)}
\end{equation}

To assess the impact of financial experts on the quality of the annotations, we plotted $vote(l)$ (Eq. \ref{voteeq}) before and after their involvement. Figure \ref{fig:ass} (i) and Figure \ref{fig:ass} (ii) show the results before and after expert checking, respectively. We observed that, initially, $11.01\%$ of the instances had a confidence value of less than 0.5. However, after expert checking, this percentage decreased to $3.49\%$. As a result, the average confidence score across all instances improved from 0.33 to 0.46. These findings suggest that financial experts significantly improved the quality of the annotations.


\subsection{Annotation Agreement}

\begin{figure}[h!]
    \centering
    \includegraphics[width=0.7\columnwidth]{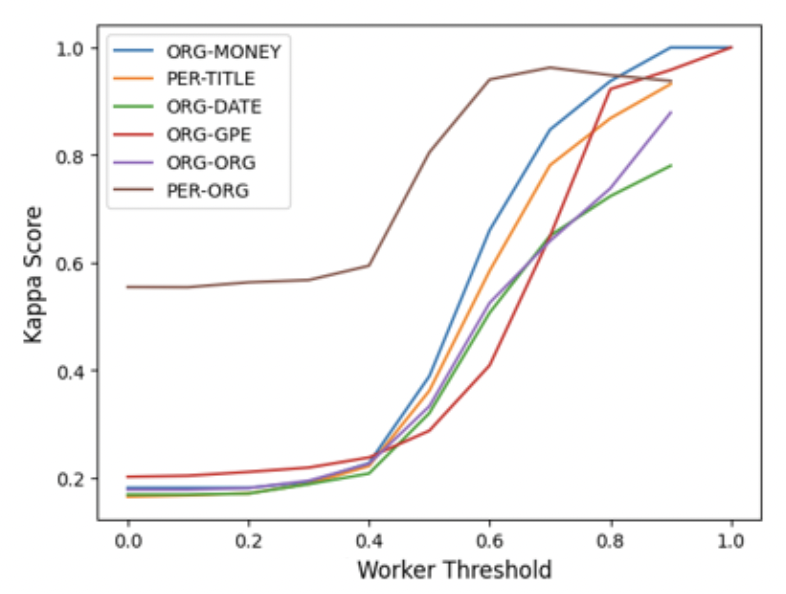}
    \caption{\textbf{Fleiss Kappa Score}. Inter-annotator agreement for each entity pair groups.} 
    \label{fig:kappa}
\end{figure}

To determine the inter-annotator agreement among the MTurk assessors, we utilized Fleiss Kappa Score \cite{Fleiss1971MeasuringNS}. In order to investigate the impact of worker reliability on our results, we conducted additional analyses by varying the trust score threshold $t$ from 0 to 1. Specifically, we excluded workers with trust scores below $t$ and recalculated the kappa scores. As depicted in Figure \ref{fig:kappa}, the Kappa score for each entity pair group varies with respect to the worker trust score threshold. Notably, we observe that removing assessments from workers with low trust scores leads to an increase in the Kappa score, suggesting that less reliable workers had a negative impact on the overall agreement among assessors. Hence, excluding less reliable workers improved the overall quality of our annotations and increased inter-annotator agreement.

It is worth noting that, for this analysis, we have merged the entity groups \textsc{per-org}, \textsc{per-univ}, and \textsc{per-gov} into a single entity group \textsc{per-org}, given that \textsc{univ} and \textsc{gov} can be regarded as specific types of organizations.

\subsection{Noise Rate}
We conducted a noise rate analysis on our dataset by randomly selecting a 5\% sample and verifying the annotations with the help of a linguist and a financial expert. The analysis revealed an overall noise rate of 6\%. We found that the primary source of noise was the \textsc{org-org} and \textsc{org-money} entity groups, which contain finance-specific relations such as \textit{acquired\_by}, \textit{shares\_of}, \textit{revenue\_of}, and \textit{cost\_of}. Annotating these groups is challenging and requires domain expertise to understand and correctly annotate these relations. We leveraged this information to enhance the quality of our dataset by correcting the errors with the help of financial experts. We have included the noise rates for each entity group in Table \ref{tab:noiserate}. 

\begin{table}[h!]
\centering
\begin{tabular}{|l|c|}
\hline
\textbf{Entity Pair} & \textbf{Noise Rate (\%)} \\ \hline
ORG:ORG               &                14      \\
ORG:GPE               &                4      \\
ORG:MONEY             &                13      \\
ORG:DATE              &                1      \\
PER:ORG               &                5      \\
PER:UNIV              &                8      \\
PER:GOV\_AGY          &                8      \\
PER:TITLE             &                3      \\ \hline
\end{tabular}
\caption{\textbf{Noise Rate}}
    \label{tab:noiserate}
\end{table}

\section{Benchmark Experiments and Results}
\label{sec:benchmark_exp}

In this section, we fine-tune various state-of-the-art deep learning models on the REFinD dataset for the relation extraction task, in order to assess and highlight the challenges posed by REFinD. To ensure a comprehensive evaluation of the benchmarks, we report both micro- and macro-F1 score metrics (evaluation details in Appendix \ref{eval}). We also evaluate the performance of each model on each entity pair group and the entire REFinD dataset to obtain a more detailed understanding of the models' strengths and weaknesses.

\begin{figure*}[tp]
    \centering
\includegraphics[width=16cm,height=6cm]{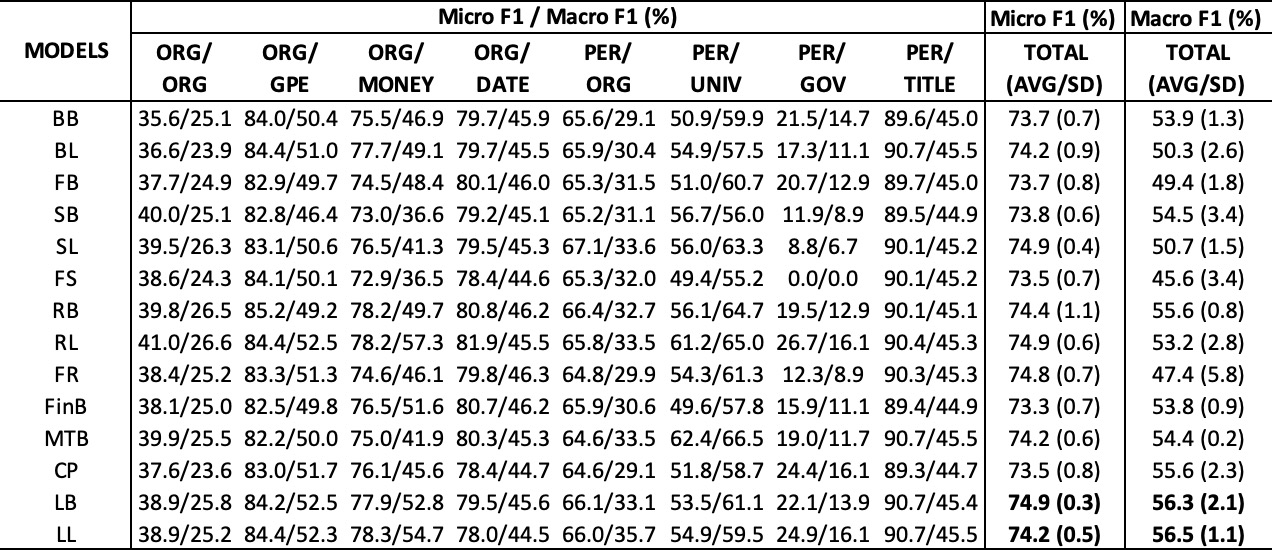}
    \caption{\textbf{REFinD Baselines.} Results achieved by the benchmark models. Reported are the
\% averages (AVG) and standard deviation (SD) over five random seeds. Here the models correspond to (i) BERT-base (BB) \cite{devlin-etal-2019-bert}, (ii) BERT-large (BL) \cite{devlin-etal-2019-bert}, (iii) FLANG-BERT (FB) \cite{shah-etal-2022-flue}, (iv) SpanBERT-base (SB) \cite{joshi2020spanbert}, (v) SpanBERT-large (SL) \cite{joshi2020spanbert}, (vi) FLANG-SpanBERT (FS) \cite{shah-etal-2022-flue}, (vii) Roberta-base (RB) \cite{liu2019roberta}, (viii) Roberta-large (RL) \cite{liu2019roberta}, (ix) FLANG-Roberta (FR) \cite{shah-etal-2022-flue}, (x) FinBERT (FinB) \cite{https://doi.org/10.48550/arxiv.1908.10063}, (xi) Matching the Blanks (MTB) \cite{soares2019matching}, (xii) Contrastive Pre-training (CP) \cite{peng2020learning}, (xiii) Luke-base (LB) \cite{yamada2020luke}, and (xiv) Luke-large (LL) \cite{yamada2020luke}.}
    \label{fig:model_comparison}
\end{figure*}

\begin{table}[h!]
\centering
\begin{tabular}{l|l}
\hline
\textbf{Parameter} & \multicolumn{1}{c}{\textbf{Value}} \\ \hline
Classifier         & 1-layer FFNN                       \\
Loss               & Cross Entropy                      \\
Optimizer          & Adam optimizer                     \\
Learning rate      & 2e-5                               \\
Batch Size         & 32                                 \\
Epochs             & 5                                  \\ \hline
\end{tabular}
\caption{\textbf{Hyperparameters Setting}. Model details for reproducibility of the baselines.} 
\label{tab:hyperparameters}
\end{table}

\subsection{Experimental Setup}
To perform relation extraction on our REFinD dataset, we adopt the architecture of Matching the Blanks \cite{soares2019matching}. First, we augment four entity markers, namely $[E1]$, $[/E1]$, $[E2]$, and $[/E2]$, to mark the beginning and end of each entity mention in a given relation instance $s$ and ordered pair of entity mentions $(e_1, e_2)$. We then use a linear classifier based on the concatenated representation of the final hidden states corresponding to the start tokens of the two entities $[E1]$ and $[E2]$ to solve the task. For fine-tuning, we use hyperparameters as outlined in Table \ref{tab:hyperparameters}.

\subsection{Benchmark Models} \label{bm}

We evaluate our REFinD dataset using various pre-trained encoders from HuggingFace\footnote{https://huggingface.co/}, including BERT-base and -large \cite{devlin-etal-2019-bert}, Roberta-base and -large \cite{liu2019roberta}, and Spanbert-base and -large \cite{joshi2020spanbert}, which have been pre-trained primarily on web-based articles. We also evaluate models that have been further trained on financial text, including FinBERT \cite{https://doi.org/10.48550/arxiv.1908.10063} and Flang encoders, namely Flang-BERT, -SpanBERT, and -Roberta \cite{shah-etal-2022-flue}. Additionally, we assess the performance of several state-of-the-art models that have been specifically pre-trained for relation extraction, such as Matching the Blanks (MTB) \cite{soares2019matching}, Contrastive Pre-training (CP) \cite{peng2020learning}, and Luke-base and -large \cite{yamada2020luke}, as benchmarks for our REFinD dataset.

\subsection{Results}

The evaluation results of the benchmark models mentioned in Section \ref{bm} are presented in Figure \ref{fig:model_comparison}. Among the benchmarks, Luke models exhibit the best overall performance, in terms of both micro- and macro-F1 scores. We also observe that entity groups like \textsc{org-org}, \textsc{per-gov}, and \textsc{per-org}, which are more prevalent in the finance domain, present greater challenges than other entity groups. Specifically, finance-domain-specific relations such as \textit{shares\_of}, \textit{cost\_of}, and \textit{member\_of} (see Figure \ref{fig:model_comparison}) exhibit F1 scores lower than 30\%, whereas general relations such as \textit{title}, \textit{headquartered\_in}, and \textit{employee\_of} achieve F1 scores greater than 70\%. This could be attributed to the fact that models such as BERT, MTB, and Luke have primarily been pre-trained on web-based articles, which exhibit a different data distribution compared to financial datasets like REFinD. Furthermore, models like FinBERT and FLANG, which have undergone additional training on financial news articles and perform masking on financial terms, still fail to achieve better results, as they have never been exposed to semantically complex finance-specific documents. Thus, we must consider these experiments as starting points, and further improvements in the financial relation extraction task are likely to be achieved through increasing model capacity and architectural innovations.

\subsection{Error Analysis}

\begin{figure}[t!]
    \includegraphics[width=\columnwidth]{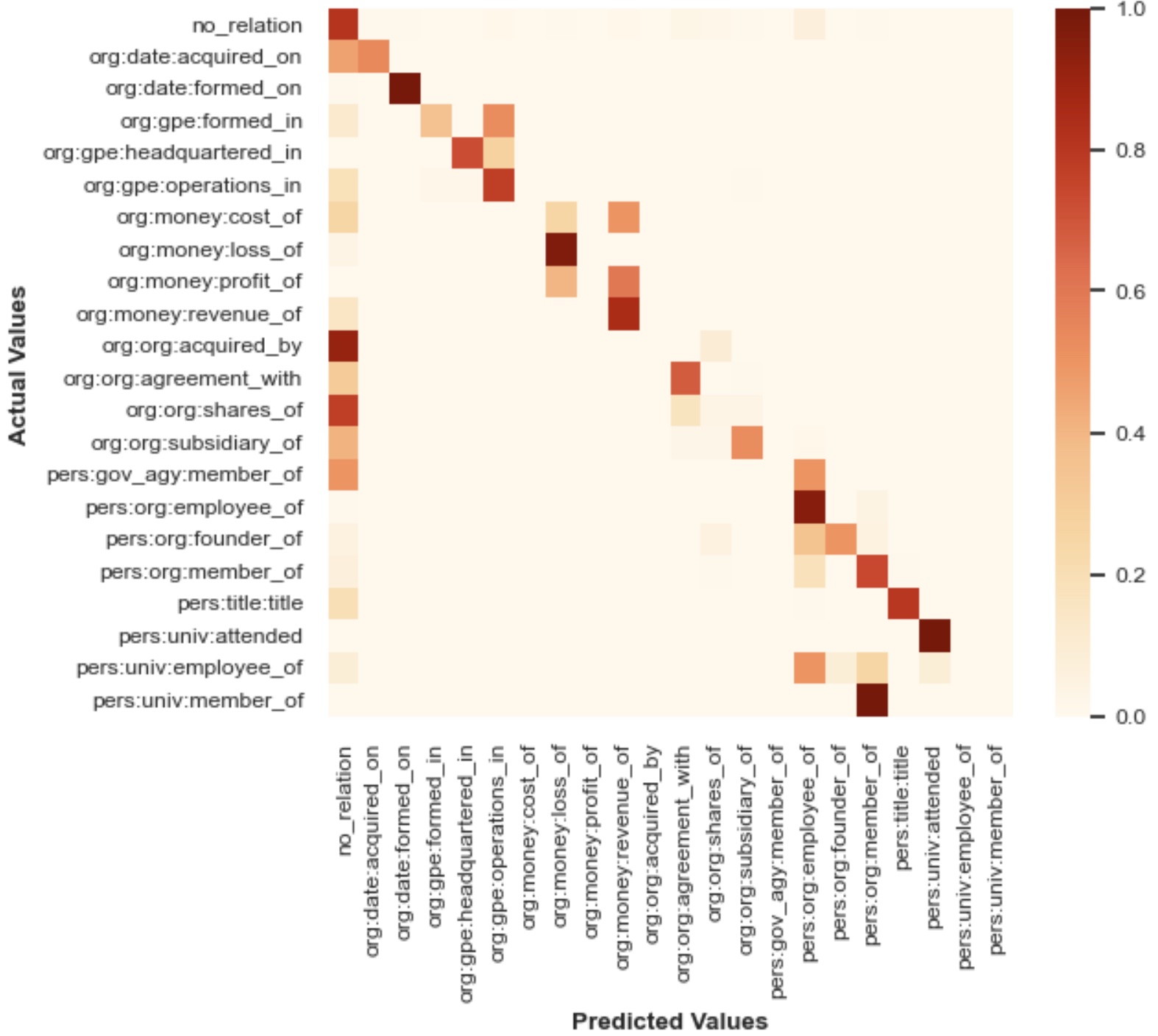} 
    \caption{Confusion Matrix for each relation in REFinD based on predictions (normalized) over Luke-large encoder.}
    \label{fig:confusionmatrix} 
\end{figure}

To provide insights for further improvements, we analyze the errors in the predictions of the relation extraction task, as shown in Figure \ref{fig:confusionmatrix}. The most common types of errors made by the models can be categorized into three groups:
\\
\noindent
\textbf{Numerical Inference:} Most current LLM models treat numbers in text in the same way as other tokens, resulting in the most common error being the failure to capture numeracy. For example, the models predict \textit{acquired\_by} for both of these instances: \textit{[ENT2]Company A[/ENT2] owns 31\% equity interest in [ENT1]Company B[/ENT1]} and \textit{[ENT1]Company A[/ENT1] has 100\% equity interest in [ENT2]Compan-y B[/ENT2]}. However, the acquisition has only occurred in the second instance.

\noindent
\textbf{Semantic Ambiguity:} Another challenge was that the models also struggled with distinguishing between similar relation types, such as \textit{member\_of}, \textit{employee\_of}, and \textit{founder\_of}, for entity pairs belonging to the \textsc{PER-ORG}, \textsc{PER-UNIV}, and \textsc{PER-GOV} groups. For example, the models often predicted \textit{employee\_of} for both of these instances: \textit{[ENT1]Jane Doe[/ENT1] is the CEO of [ENT2]Company B[/ENT2]} and \textit{[ENT1]John Doe[/ENT1] is on the Board of Directors of [ENT2]Company B[/ENT2]}. However, in the second instance, John Doe is only a member of Company B, not an employee. This suggests that the models may need to learn more nuanced distinctions between similar relation types, which could be addressed through improvements in training data, model architecture, or both.

\noindent
\textbf{Directional Ambiguity:} The models exhibit uncertainty in determining the directional dependency between two entities for the relations \textit{acquired\_by} and \textit{subsidiary\_of}, despite being provided with the order of the entity pairs. This leads to errors where the models predict \textit{subsidiary\_of} for both instances in the phrases "Company A is the subsidiary of Company B" and "After purchasing Company A, through its wholly owned subsidiary, Company C purchased Company B". In reality, the second instance implies that Company B was acquired by Company A. 

\section{Ethics Statement}

This work relies on the use of documents obtained from the SEC website.  SEC filings are freely and publicly available, but do contain names and other identifying details such as current and former job titles, schools attended, and board or professional associations for typically high-ranking officers in publicly traded companies.  This information is similar to what could be found about public figures mentioned in websites such as Wikipedia and thus we do not anticipate any harm to persons mentioned in our data beyond what could be learned from reading the public financial statements themselves.  Additionally, we have made no attempt to aggregate or include non-public information about any individual or entity in this dataset.  All data used for dataset construction was intentionally selected to be several years old as of this publication and thus we do not anticipate any impact on financial markets with this release. REFinD is released under a license for non-commercial use. 

\section{Conclusion and Future Work}
\label{conclusion}

In this paper, we introduced REFinD, which is the largest-scale annotated dataset of relations generated entirely over financial documents to-date, aimed to support the development of downstream applications within the finance domain. We highlighted the challenges involved in collecting such a large-scale relation extraction dataset specifically over financial text and emphasized the importance of financial-domain expertise in annotating such datasets. Furthermore, we fine-tuned various state-of-the-art deep learning models on the REFinD dataset to identify the challenges posed by our dataset. Our results showed that these models do not perform well on finance-specific relations, mainly because they have not been exposed to complex financial text and documents. We also identified three main challenges involved in relation extraction over financial text, namely numeracy, ambiguity amongst relations, and the direction of relations.

As for future directions, we plan to enhance the dataset by adding more types of finance-specific entity groups and hence more relations. Moreover, we aim to improve the dataset to capture the comparison between financial events over time, which is crucial in financial text analysis. For instance, we plan to include examples that capture the comparison between revenues earned by a company over different years, such as \it Software LLC earned a net revenue of \$1.5 billion in 2018 as compared to \$1 billion in 2017\rm. These improvements will help us address the challenges involved in relation extraction over financial text and enable the development of more accurate and effective financial text analysis tools.

\section{Acknowledgements}
\label{acknowledgements}

We would like to thank Natraj Raman, Daniel Borrajo, Armineh Nourbakhsh, Elena Kochkina, Zhiqiang Ma, and our anonymous reviewers for their thoughtful comments and feedback which greatly contributed to the quality of this work.

Disclaimer. This paper was prepared for informational purposes by the Artificial Intelligence Research group of JPMorgan Chase \& Co. and its affiliates (“JP Morgan”), and is not a product of the Research Department of JP Morgan. JP Morgan makes no representation and warranty whatsoever and disclaims all liability, for the completeness, accuracy or reliability of the information contained herein. This document is not intended as investment research or investment advice, or a recommendation, offer or solicitation for the purchase or sale of any security, financial instrument, financial product or service, or to be used in any way for evaluating the merits of participating in any transaction, and shall not constitute a solicitation under any jurisdiction or to any person, if such solicitation under such jurisdiction or to such person would be unlawful.

\appendix

\section{Appendix}
\label{sec:appendix}

\begin{figure*}[tbp]
    \centering
    \includegraphics[width=0.5\textwidth]{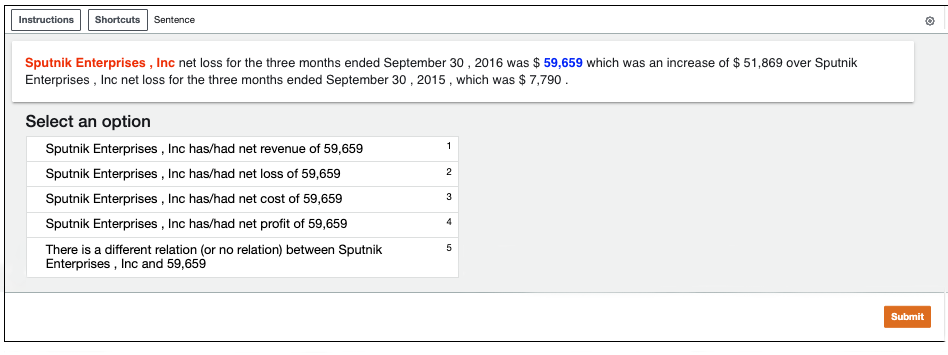}
    \caption{Select the statement that best describes the relation in the example sentence.}
    \label{fig:mturk}
\end{figure*}

\subsection{MTurk Annotation Example} \label{mturkeg}
The entity pair ($e_1$,$e_2$) represents one instance.  One sentence can have multiple entity pairs.  Hence, multiple instances could be labeled from the same sentence.  For example, in Figure \ref{fig:mturk}, we see a sample HIT with the highlighted entities and options to be selected below.  Because there are two mentions of \textit{Sputnik Enterprises, Inc} and three currency amounts, we have 6 possible combinations of the relation \textsc{org-money}.  To avoid ambiguity, the red and blue highlights indicate which entities in the sentence are in focus.

\subsection{Dataset Construction} \label{app:datcon}

The Hit score can be defined as the equation below:
\begin{equation}
\scriptsize
\textit{Hit}(\textit{p}\vert X,S) = \sum_{x_i\in X}\sum_{s_j\in S}[match(x_i,p,s_j)]
\label{hit-score}
\end{equation}

Given an entity pair $x=(e_1,e_2)$, and sentence $s in S$, the \textit{Hit} score sums the frequency of candidate seed patterns matching $x$.

The Conv score can be defined as:
\begin{equation}
\textit{Conv}(\textit{p}\vert X,S) = \frac{\sum_{x_i\in X}[\sum_{s_j\in S}[match(x_i,p,s_j)]>0]}{|X|}
\label{conv-score}
\end{equation}



\subsection{Evaluation} \label{eval}
Since the REFinD dataset comprises of 45.5\% of \textsc{no\_relation}, we adopt the strict f1-metric so as to accurately measure performance on remaining relations. More specifically, while accounting for the instances used to calculate the f1-score, we first remove the instances where the model correctly predicts \textsc{no\_relation}. For the below remaining instances, we calculate the f1-score and compare the performance, (i) instances where both true and predicted relations are not equal to \textsc{no\_relation}, (ii) instances where true relation is \textsc{no\_relation} and predicted relation is not equal to \textsc{no\_relation}, (iii) instances where true relation is not equal to \textsc{no\_relation} and predicted relation is equal to \textsc{no\_relation}.


\subsection{REFinD Snippet} \label{app:datasnip}

\{\textbf{"id"}: "10Q\_edgar\_data\_1408710\_0001193125-17033383\_1.txt",
 \textbf{"relation"}: "org:gpe:operations\_in",
 \textbf{"rel\_group"}: "ORG-GPE",
\textbf{"token"}: ["For", "instance", ",", "Fabrinet", "have", "intercompany", "agreements", "in", "place", "that", "provide", "for", "Fabrinet", "California", "and", "Singapore", "subsidiaries", "to", "provide", "administrative", "services", "for", "the", "Cayman", "Islands", "Parent", ", ", "and", "the", "Cayman", "Islands", "Parent", "has", "entered", "into", "manufacturing", "agreements", "with", "Fabrinet", "Thai", "subsidiary", "."],
 \textbf{"e1\_start"}: 12, \textbf{"e1\_end"}: 13, \textbf{"e2\_start"}: 23, \textbf{"e2\_end"}: 25, \textbf{"e1\_type"}: "ORG", \textbf{"e2\_type"}: "GPE",
\textbf{"spacy\_pos"}: ["IN", "NN", ", ", "NNP", "VBP", "JJ", "NNS", "IN", "NN", "WDT", "VBP", "IN", "NNP", "NNP", "CC", "NNP", "NNS", "TO", "VB", "JJ", "NNS", "IN", "DT", "NNP", "NNP", "NN", ", ", "CC", "DT", "NNP", "NNP", "NNP", "VBZ", "VBN", "IN", "NN", "NNS", "IN", "NNP", "NNP", "NN", "."],
 \textbf{"spacy\_ner"}: ["O", "O", "O", "O", "O", "O", "O", "O", "O", "O", "O", "O", "ORG", "O", "O", "O", "O", "O", "O", "O", "O", "O", "O", "GPE", "GPE", "O", "O", "O", "O", "O", "O", "O", "O", "O", "O", "O", "O", "O", "O", "O", "O", "O"],
 \textbf{"spacy\_head"}: [4, 0, 4, 4, 4, 6, 4, 4, 7, 10, 6, 18, 16, 16, 13, 13, 18, 18, 10, 20, 18, 18, 25, 24, 25, 21, 4, 4,31, 30, 31, 33, 33, 4, 33, 36, 34, 33, 39, 40, 37, 33],
 \textbf{"spacy\_deprel"}:["prep", "pobj", "punct", "nsubj", "ROOT", "amod", "dobj", "prep", "pobj", "nsubj", "relcl", "mark","nmod", "nmod", "cc", "conj", "nsubj", "aux", "advcl", "amod", "dobj", "prep", "det", "compound", "compound", "pobj", "punct", "cc", "det", "compound", "compound", "nsubj", "aux", "conj", "prep", "compound", "pobj", "prep", "compound", "compound", "pobj", "punct"],
 \textbf{"sdp"}: ["Fabrinet", "California", "provide", "provide", "services", "Parent", "Cayman Islands"], \textbf{"sdp\_tok\_idx"}: [13,10,18,20,25]\}

\subsection{REFinD Statistical Breakdown by Individual Relation}
\label{refinddatastats}

\begin{table}[h!]
\begin{tabular}{lllll}
\hline
Relation                & \#Train & \#Dev & \#Test & \#Total \\
\hline
no/other\_relation & 9128 & 1965 & 1953 & 13046 \\
pers:title:title & 3126 & 671 & 671 & 4468 \\
org:gpe:operations\_in & 2832 & 606 & 605 & 4043 \\
pers:org:employee\_of & 1733 & 372 & 374 & 2479 \\
org:org:agreement\_with & 653 & 141 & 141 & 935 \\
org:date:formed\_on & 448 & 96 & 96 & 640 \\
pers:org:member\_of & 441 & 94 & 95 & 630 \\
org:org:subsidiary\_of & 386 & 82 & 83 & 551 \\
org:org:shares\_of & 286 & 61 & 61 & 408 \\
org:money:revenue\_of & 217 & 47 & 47 & 311 \\
org:money:loss\_of & 141 & 30 & 31 & 202 \\
org:gpe:headquartered\_in & 135 & 29 & 29 & 193 \\
org:date:acquired\_on & 134 & 28 & 24 & 186 \\
pers:org:founder\_of & 92 & 19 & 20 & 131 \\
org:gpe:formed\_in & 81 & 17 & 17 & 115 \\
org:org:acquired\_by & 55 & 11 & 12 & 78 \\
pers:univ:employee\_of & 53 & 11 & 12 & 76 \\
pers:gov\_agy:member\_of & 40 & 8 & 8 & 56 \\
pers:univ:attended & 30 & 6 & 7 & 43 \\
pers:univ:member\_of & 23 & 5 & 5 & 33 \\
org:money:profit\_of & 20 & 4 & 5 & 29 \\
org:money:cost\_of & 16 & 3 & 4 & 23 \\
\hline
\#Total & 20070 & 4306 & 4300 & 28676 \\
\hline
\end{tabular}
\caption{REFinD statistical breakdown (i.e., number of instances) by individual relation.}
\label{tab:relation_breakdown}
\end{table}

\bibliographystyle{ACM-Reference-Format}
\balance
\bibliography{REFinD_sigconf}

\end{document}